\pdfoutput=1
\documentclass[11pt]{article}

\usepackage[]{naacl2021}

\usepackage{times}
\usepackage{latexsym}

\usepackage[T1]{fontenc}
\usepackage[utf8]{inputenc}

\usepackage{microtype}

\usepackage{url}
\usepackage{xcolor}
\usepackage{booktabs}
\usepackage[T1]{fontenc}

\usepackage{graphicx}
\usepackage{xspace}
\graphicspath{ {./figures/} }
\usepackage{amsmath}
\usepackage{cleveref}
\usepackage[group-separator={,}]{siunitx}
\usepackage{ifthen}
\usepackage{footnote}

\usepackage{siunitx}
\makeatletter
\def\blfootnote{\xdef\@thefnmark{}\@footnotetext}
\makeatother

\title{
\sword \swordemoji: A Benchmark for Lexical Substitution \\with Improved Data Coverage and Quality
}

\author{
Mina Lee\textsuperscript{*} \And Chris Donahue\textsuperscript{*} \And Robin Jia \\
Stanford University \\
\texttt{\{minalee, cdonahue, robinjia, aiyabor, pliang\}@cs.stanford.edu} \\
\And Alexander Iyabor \And Percy Liang
}

\newif\ifcomment
\commenttrue
\ifcomment
\newcommand{\todo}[1]{{\color{blue} TODO: {#1}}}
\newcommand{\ml}[1]{{\color{purple} [M] {#1}}}
\newcommand{\cd}[1]{{\color{red} [C] {#1}}}
\newcommand{\rj}[1]{{\color{red} [R] {#1}}}
\newcommand{\pl}[1]{{\color{red} [P] {#1}}}
\else
\newcommand{\todo}[1]{}
\newcommand{\ml}[1]{}
\newcommand{\rj}[1]{}
\newcommand{\cd}[1]{}
\newcommand{\pl}[1]{}
\fi

\newcommand{\semeval}[0]{\textsc{SemEval}\xspace}
\newcommand{\coinco}[0]{\textsc{CoInCo}\xspace}
\newcommand{\sword}[0]{\textsc{Swords}\xspace}

\newcommand{\coincoonly}[0]{\coinco-only\xspace}
\newcommand{\bothcoincoandthesaurus}[0]{\coinco $\cap$ Thesaurus\xspace}
\newcommand{\thesaurusonly}[0]{Thesaurus-only\xspace}

\newcommand{\coverage}[0]{4.1x\xspace}
\newcommand{\quality}[0]{1.5x\xspace}

\newcommand{\pk}{$P_c^{k}$}
\newcommand{\rk}{$R_c^{k}$}
\newcommand{\fk}{$F_c^{k}$}
\newcommand{\pmk}{$P^{k}$}
\newcommand{\rmk}{$R^{k}$}
\newcommand{\fmk}{$F^{k}$}

\newcommand{\pten}{$P_c^{10}$}
\newcommand{\pmten}{$P^{10}$}
\newcommand{\rten}{$R_c^{10}$}
\newcommand{\rmten}{$R^{10}$}
\newcommand{\ften}{$F_c^{10}$}
\newcommand{\fmten}{$F^{10}$}
\newcommand{\Gstrict}{Strict}

\newcommand{\Gperm}{Lenient}
\newcommand{\gperm}{lenient}

\newcommand{\baformat}[1]{\textsc{#1}}
\newcommand{\baoracle}{\baformat{Oracle}}
\newcommand{\bahumans}{\baformat{Humans}}
\newcommand{\bacoinco}{\baformat{CoInCo}}
\newcommand{\babertk}{\text{\baformat{Bert-K}}}
\newcommand{\babertm}{\text{\baformat{Bert-M}}}
\newcommand{\babbls}{\text{\baformat{Bert-LS}}}
\newcommand{\bagpt}{\text{\baformat{Gpt-3}}} % added \text to prevent the word from being broken into two lines 
\newcommand{\bawordtune}{\baformat{Wordtune}}
\newcommand{\bathesaurus}{\baformat{Thesaurus}}
\newcommand{\babert}{\baformat{Bert}}
\newcommand{\baglove}{\baformat{GloVe}}
\newcommand{\barandom}{\baformat{Random}}

\newcommand\swordemoji{\raisebox{-2pt}{\includegraphics[width=0.9em]{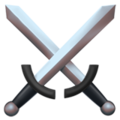}}\xspace}

\begin{document}
\maketitle
\begin{abstract}
We release a new benchmark for lexical substitution, the task of finding appropriate substitutes for a target word in a context.
To assist humans with writing, 
lexical substitution systems can suggest words that humans cannot easily think of.
However, existing benchmarks depend on human recall as the only source of data,
and therefore lack coverage of the substitutes that would be most helpful to humans.
Furthermore, annotators often provide substitutes of low quality, which are not actually appropriate in the given context.
We collect higher-coverage and higher-quality data by framing lexical substitution as a classification problem, guided by the intuition that it is easier for humans to judge the appropriateness of candidate substitutes than conjure them from memory. 
To this end, we use a context-free thesaurus to produce candidates and rely on human judgement to determine contextual appropriateness. 
Compared to the previous largest benchmark, our \sword benchmark has \coverage more substitutes per target word for the same level of quality, 
and its substitutes are \quality more appropriate (based on human judgement) for the same number of substitutes.
\end{abstract}

\blfootnote{\textsuperscript{*}Equal contribution.}

\section{Introduction}
Imagine you are writing the message ``I read an amazing paper today'' to a colleague, 
but you want to choose a more descriptive adjective to replace ``amazing.''
At first you might think of substitutes like ``awesome'' and ``great,'' but feel that these are also unsatisfactory. 
You turn to a thesaurus for inspiration, but among reasonable alternatives like ``incredible'' and ``fascinating'' are words like ``prodigious'' which do not quite fit in your context. 
Ultimately, you choose to go with ``fascinating,'' but reaching this decision required a non-trivial amount of time and effort. 

\looseness-1 Research on \emph{lexical substitution}~\citep{mccarthy-2002-lexical,mccarthy-navigli-2007-semeval,erk-pado-2008-structured,szarvas-etal-2013-learning,kremer-etal-2014-substitutes,melamud-etal-2015-modeling,hintz-biemann-2016-language,zhou-etal-2019-bert,arefyev-etal-2020-comparative} 
considers the task of replacing a target word in context with appropriate substitutes. 
There are two widely-used English benchmarks for this task: \semeval~\citep{mccarthy-navigli-2007-semeval} and \coinco~\citep{kremer-etal-2014-substitutes}.
For both benchmarks, data was collected by asking human annotators to think of substitutes from memory. 
Because lexical substitution was originally proposed as a means for evaluating word sense disambiguation systems~\citep{mccarthy-2002-lexical}, 
this data collection strategy was designed to avoid a bias towards any particular word sense inventory.

\looseness-1 In this work, we consider a different use case for lexical substitution:
writing assistance.
For this use case, we are interested in evaluating a system's ability to produce appropriate substitutes that are likely to be difficult for humans to think of.
We show that the data collection strategy used in past benchmarks yields low \emph{coverage} of such uncommon substitutes---for our previous example, they might contain words like ``awesome'' and ``great,'' but miss words like ``incredible'' and ``fascinating.''
Furthermore, we observe that these benchmarks have low \emph{quality}, containing words like ``fun,'' which are easy to think of, but not quite appropriate in context.

\begin{table*}[t]
\centering
\small
\begin{tabular}{p{0.09\linewidth}p{0.85\linewidth}}
    \toprule
    Context & My favorite thing about her is her \underline{straightforward} honesty. \\
    \midrule
    \coinco & candid~(3), artless~(1), blunt~(1), complete~(1), direct~(1), forthright~(1), frank~(1), outspoken~(1), plainspoken~(1), truthful~(1), unreserved~(1) \\
    \addlinespace[1mm]
    Thesaurus & aboveboard, apparent, barefaced, candid, clear, clear-cut, direct, distinct, easy, elementary, evident, forthright, frank, genuine, guileless, honest, honorable, just, laid on the line, level, like it is, manifest, ... 32 more \\
    \addlinespace[1mm]
    \sword & \textbf{sincere}~(80\%), \textbf{genuine}~(80\%), \textbf{frank}~(70\%), \textbf{candid}~(70\%), \textbf{direct}~(70\%), \textbf{forthright}~(70\%), \textbf{uncomplicated}~(60\%), \textbf{up front}~(60\%), \textbf{clear-cut}~(60\%), \textbf{clear}~(60\%), \textbf{plainspoken}~(60\%), complete~(50\%), straight-arrow~(50\%), honest~(50\%), open~(50\%), blunt~(50\%), outspoken~(50\%), truthful~(50\%), plain-dealing~(40\%), undisguised~(40\%), unvarnished~(40\%), unreserved~(40\%), barefaced~(40\%), unequivocal~(30\%), upright~(30\%), simple~(30\%), veracious~(30\%), unconcealed~(30\%), like it is~(30\%), square-shooting~(20\%), upstanding~(20\%), undissembled~(20\%), manifest~(20\%), unambiguous~(20\%), pretenseless~(20\%), level~(10\%), laid on the line~(10\%), honorable~(10\%), guileless~(10\%), ... 20 more with 0\% \\
    \bottomrule
\end{tabular}
\caption{
We consider lexical substitution, the task of finding appropriate substitutes for a \underline{target word} in context. 
In \coinco (the previous largest benchmark), humans are asked to think of substitutes from memory and result in low coverage
(the number of annotators who produced each substitute is shown in parentheses; out of six annotators). 
On the other hand, looking up the target word in a thesaurus has higher coverage, but low quality, because it does not consider the context.
In \sword, we combine the best of both worlds and provide a list of substitutes that has high coverage and high quality, along with fine-grained scores for each substitute (shown in parentheses). 
Substitutes with scores greater than 50\% from \sword are \textbf{bolded}.
}
\label{tab:example}
\end{table*}

We present \sword---the \underline{S}tanford \underline{Word} \underline{S}ubstitution Benchmark---an English lexical substitution benchmark that raises the bar for both coverage and quality (\Cref{tab:example}).
We collect \sword by asking human annotators to \emph{judge} whether a given candidate word is an appropriate substitute for a target word in context, following the intuition that judging a given substitute is easier than producing that same substitute from memory.
To bootstrap a set of candidates for humans to annotate, we look up target words in an existing context-free thesaurus \citep{kipfer-2013-roget}.\footnote{Note that our use of a thesaurus makes \sword{} less appropriate for the original use case for lexical substitution: evaluating word sense disambiguation systems.}
Because a thesaurus might miss substitutes that would not typically be synonymous with the target word outside of the provided context (e.g. ``thought-provoking'' for ``amazing''), we also include human-proposed candidates from the previous \coinco{} benchmark.

Determining whether a substitute is appropriate is intrinsically subjective. 
To address this, we collect binary labels from up to ten annotators for each substitute, inducing a score for each substitute. 
In \coinco, analogous scores are derived from the number of independent annotators who thought of a substitute---hence, as we will show in~\Cref{sec:data_analysis}, these scores tend to correspond more to ease-of-recollection than appropriateness. 
In contrast, scores from \sword{} correspond to appropriateness, and also allow us to explicitly trade off coverage and quality, permitting more nuanced evaluation.
Our analysis shows that compared to \coinco, \sword has \coverage more substitutes per target word for the same level of quality, and its substitutes are \quality more appropriate based on scores for the same number of substitutes.

We demonstrate that \sword is a challenging benchmark by evaluating state-of-the-art lexical substitution systems and large-scale, pre-trained language models including systems based on BERT \citep{devlin-etal-2019-bert,zhou-etal-2019-bert} and \text{GPT-3} \citep{brown-etal-2020-gpt3}. 
In our evaluation, we find that humans substantially outperform all existing systems, 
suggesting that lexical substitution can be used as a downstream language understanding task for pre-trained models.
We release \sword publicly as a benchmark for lexical substitution,
coupled with a Python library that includes previous benchmarks in a common format, 
standardized evaluation scripts for prescribed metrics, 
and reproducible re-implementations of several baselines.\footnote{
\sword: \href{https://github.com/p-lambda/swords}{\url{github.com/p-lambda/swords}} \\
All experiments reproducible on the CodaLab platform: \\ \href{https://worksheets.codalab.org/worksheets/0xc924392d555f4b4fbee47be92e3daa0b}{\url{worksheets.codalab.org/worksheets/0xc924392d555f4b4fbee47be92e3daa0b}}
}

 \section{Background}
\label{sec:background}
\begin{table*}[t]
\centering
\small
\vspace{-0.2cm}
\begin{tabular}{
        rrrrr
        S[table-format=2.1]
        S[table-format=2.1]
        S[table-format=2.1]
        S[table-format=1.1]
    }
    \toprule
    Benchmark & Contexts & Targets & Substitutes & Labels & \multicolumn{4}{c}{Substitutes per target (on average)} \\
    \cmidrule(lr){6-9}
    & & & & & {Total} & {Inconceivable} & {Conceivable} & {Acceptable} \\
    & & & & & & {(score $= 0$\%)} & {(score $> 0$\%)} & {(score $> 50$\%)} \\
    \midrule
    \semeval & \num{2010} & \num{201} & \num{8025} & \num{12300} & 4.0 & {-} & {-} & {-} \\
    \coinco  & \num{2474} & \num{15629} & \num{112742} & \num{167446} & 7.2 & 2.5* & 5.2* & 2.3* \\
    \sword  & \num{1132} & \num{1132} & \num{68683} & \num{375855} & 60.7 & 39.3 & 21.4 & 4.1 \\
    \hline
    \addlinespace[1mm]
    \coinco (dev) & \num{1577} & \num{10179} & \num{67814} & \num{98950} & 6.7 & 2.2* & 5.2* & 2.5* \\
    \sword (dev) & \num{370} & \num{370} & \num{22978} & \num{121938} & 62.1 & 41.6 & 20.5 & 4.2 \\
    \hline
    \addlinespace[1mm]
    \coinco (test) & \num{897} & \num{5450} & \num{44928} & \num{68496} & 8.2 & 2.9* & 5.7* & 2.4* \\
    \sword (test) & \num{762} & \num{762} & \num{45705} & \num{253917} & 60.0 & 38.1 & 21.9 & 4.0 \\
    \bottomrule
\end{tabular}
\caption{Benchmark statistics for \semeval, \coinco, and \sword. 
Our benchmark contains \coverage more conceivable substitutes (21.4) and 1.8x more acceptable substitutes (4.1) on average, compared to the previous largest benchmark \coinco.
Unlike prior benchmarks, \sword has inconceivable substitutes that received a score of~0 from appropriateness judgement, which are useful as challenging distractors for model evaluation.
The numbers of inconceivable, conceivable, and acceptable substitutes for \coinco (numbers with *) are estimated based on the subset of \coinco used to build \sword and therefore received scores.
}
\label{tab:summary}
\vspace{-0.1cm}
\end{table*}

We describe lexical substitution and briefly introduce two widely-used benchmarks: \semeval \citep{mccarthy-navigli-2007-semeval}, the first benchmark, and \coinco \citep{kremer-etal-2014-substitutes}, the largest existing benchmark.
For a survey of other benchmarks, we refer readers to \citet{kremer-etal-2014-substitutes}, \citet{hintz-biemann-2016-language}, and \citet{miller-2016-thesis}.

\paragraph{Lexical substitution.}
Lexical substitution is the task of generating a list of substitutes $w'$ that can replace a given target word $w$ in a given context~$c$ \citep{mccarthy-2002-lexical}:
\begin{equation*}
(\text{context}~c, \text{target}~w) \rightarrow [\text{substitute}~w'].
\end{equation*}
The context $c$ is one or more sentences where the target word $w$ is situated.
The target word $w$ is one word in the context, which is either manually chosen by humans~\citep{mccarthy-navigli-2007-semeval} or automatically selected based on the part-of-speech of the target word~\citep{kremer-etal-2014-substitutes}.
The substitute $w'$ can be a word or phrase.
Note that the task of lexical substitution does not consider inflection and does not involve grammar correction;
all benchmarks contain lemmas as substitutes (e.g. ``run'' instead of ``ran'').

\paragraph{\semeval.} 
The first lexical substitution benchmark, \semeval-2007 Task~10~\citep{mccarthy-navigli-2007-semeval}, contains $201$ manually chosen target words.
For each target word, $10$ sentences were chosen as contexts (mostly at random, but in part by hand) 
from the English Internet Corpus \citep{sharoff-2006-corpora} and presented to five human annotators.
The five annotators were instructed to produce up to three substitutes from memory as a replacement for the target word in context that ``preserves the meaning of the original word.''
This resulted in \num{12300} labels in total with four substitutes per target word on average.

\paragraph{\coinco.}
The previous largest lexical substitution benchmark, \coinco~\citep{kremer-etal-2014-substitutes}, was constructed by first choosing \num{2474} contexts from the Manually Annotated Sub-Corpus \citep{ide-etal-2008-masc,ide-etal-2010-manually}.
Then, all content words (nouns, verbs, adjective, and adverbs) in the sentences were selected to be target words. 
Each target word was presented to six human annotators, who were asked to provide up to five substitutions or mark it as unsubstitutable.
All the annotators were instructed to provide (preferably single-word) substitutes for the target that ``would not change the meaning.''
This resulted in \num{167446} labels in total and \num{7.2} substitutions per target word on average.\footnote{The reported number in \citet{kremer-etal-2014-substitutes} is \num{167336} and \num{10.71}, respectively. The latter differs as they counted the same substitute multiple times when suggested by multiple annotators, whereas we report the number of unique substitutes. We also find that scores under \coinco (i.e. the number of annotators who provided a substitute) could be inflated, as at least $375$ substitutes have scores greater than six (up to $21$).}
For the rest of the paper, we focus on \coinco (but not \semeval) as our benchmark is built on \coinco and it is the largest existing benchmark.

\section{Our benchmark}
\looseness-1 \sword is composed of context, target word, and substitute triples ($c, w, w'$), each of which has a score that indicates the appropriateness of the substitute.
We consider a substitute to be \textit{acceptable} if its score is greater than 50\% (e.g. bolded words in Table \ref{tab:example}) and \textit{unacceptable} if the score is less than or equal to 50\%.
Similarly, a substitute with a score greater than 0\% is considered \textit{conceivable}, and otherwise \textit{inconceivable}. 
Note that these terms are operational definitions for convenience, and different thresholds can be chosen for desired applications.

\subsection{Addressing limitations of past work}
\label{sec:addressing}

\paragraph{Improving quality.} 
In prior work, annotators were prompted to consider whether a substitute ``preserves the meaning'' \citep{mccarthy-navigli-2007-semeval} or ``would not change the meaning'' \citep{kremer-etal-2014-substitutes} of the target word.
Instead, we ask annotators whether they ``would actually consider using this substitute as the author of the original sentence.''
We believe this wording encourages a higher standard.
In Section~\ref{sec:quality}, we provide evidence that substitutes from \sword have higher quality than those from \coinco on average.

\paragraph{Improving coverage.}
For prior benchmarks, annotators were asked to generate a list of substitutes from memory.
Psycholinguistic studies have shown that when humans are asked to predict the next word of a sentence,
they deviate systematically from the true corpus probabilities \citep{smith-levy-2011-cloze,eisape-etal-2020-cloze}.
Thus, we may reasonably expect that asking humans to generate substitutes would similarly lead to systematic omissions of some appropriate substitutes.

We observe that prior benchmarks exclude many appropriate substitutes that are difficult for humans to think of (Section \ref{sec:coverage}). 
To address this limitation, we first obtain a set of candidate substitutes and then ask annotators to \textit{judge} whether they would consider using a given candidate to replace the target word in the context.
That is, given a context~$c$, target word~$w$, and candidate substitute~$w'$, we ask humans to judge whether~$w'$ is a good replacement for the target word:
\begin{equation*}
(\text{context}~c, \text{target}~w, \text{substitute}~w’) \rightarrow \{0, 1\},
\end{equation*}
where a \textit{positive} label~$1$ corresponds to ``I would actually consider using this substitute as the author of the original sentence,'' 
and a \textit{negative} label~$0$ as the opposite.
As described in Section~\ref{sec:collection}, we annotate a large pool of candidate substitutes to ensure high coverage of all possible substitutes.
We confirm that this increases coverage compared to \coinco in Section~\ref{sec:coverage}.

\paragraph{Redefining scores to reflect appropriateness.}
In past work, each substitute~$w'$ has an associated score defined as the number of annotators who produced~$w'$ given the associated context~$c$ and target word~$w$.
Instead, we define the score as the fraction of annotators who judged the~$w'$ to be an appropriate replacement of~$w$.
We argue that the previous definition of score reflects ease-of-recollection, but not necessarily appropriateness.
In Section \ref{sec:adequacy}, we show that our definition of score better represents the appropriateness of each substitute.

\subsection{Data collection}
\label{sec:collection}
We collect substitutes and scores for a context and target word pair $(c, w)$ via the following three steps.

\paragraph{Step~1: Select contexts, targets, and substitutes.}
We use the subset of contexts and target words from \coinco.
Concretely, we start with the $(c, w)$ pairs in \coinco and randomly select one~$w$ per~$c$ to annotate.
Here, the context~$c$ consists of three sentences, where the middle sentence has the target word~$w$ (see \Cref{sec:dedup_context} to see how we handled duplicate contexts from \coinco).

Next, we choose a set of candidate substitutes~$w'$ to annotate for each $(c, w)$ pair,
as framing annotation as binary classification requires determining the set of candidate substitutes a priori.
We use human-generated substitutes from \coinco, then add substitutes from an existing context-free thesaurus, Roget's Thesaurus (\citet{kipfer-2013-roget}; see \Cref{sec:app_thesaurus} for details).
In principle, candidate substitutes can be retrieved from any lexical resource or even sampled from a generative model, which we leave as future work.
By combining candidates from \coinco and the thesaurus, we increase the coverage of acceptable substitutes.

\paragraph{Step~2: Reduce the pool of substitutes.}
Given a list of candidate substitutes from the previous step, we collect three binary labels on each $(c, w, w')$ triple (see Section~\ref{sec:crowdsourcing} for details).
Then, we pass any substitute with at least one positive label to Step~3 and further collect fine-grained scores.
We show that the probability that an acceptable substitute gets incorrectly filtered out as an inconceivable substitute is very low in Section \ref{sec:additionaldata}.

\paragraph{Step~3: Collect fine-grained scores.}
In the final step, we collect seven more binary labels on the substitutes which received at least one positive label from Step~2.
This yields a total of~$10$ binary labels for the substitutes.

\subsection{Crowdsourcing}
\label{sec:crowdsourcing}
We used Amazon Mechanical Turk (AMT) to crowdsource labels on substitutes.
Each Human Intelligence Task (HIT) contained a target word highlighted in the context and at most~$10$ candidate substitutes for the target word.
Each candidate substitute had three radio buttons for positive, negative, and abstain.
Annotators were asked to choose positive if they would actually consider using the substitute to replace the target word as the author of the context, negative if they would not consider using the substitute, and abstain if they do not know the meaning of the substitute.
We treated all abstain labels ($1.2$\% of total labels) as negative labels, thereby making it binary. 
The benchmark includes abstain labels to maintain the option for them to be handled separately (e.g. excluded) in the future.
The interface, instructions, qualification conditions, and filtering criteria used for crowdsourcing can be found in Appendix~\ref{app:crowdsourcing}.

\begin{table*}[t]
\centering
\small
\vspace{-0.1cm}
\begin{tabular}{lccc}
    \toprule
    Context with \underline{target word} & Substitute & \coinco's score & \sword's score \\
    % & & (max: 6) & (max: 100\%) \\
    \midrule
    Listen, \underline{man}, he's not there! & kid & 4 & 0\% \\
    % I did not believe love had ever \underline{existed} for anyone. & prevail & 3 & 40\% \\
    We might have a \underline{lot} of work ahead of us. & bit & 2 & 0\% \\
    She was \underline{heading} for a drink and slipped out of the crowd. & look & 2 & 10\% \\
    % She seemed so depthless - self conscious and \underline{shallow} on the outside. & flat & 2 & 0\% \\
    \addlinespace[1mm]
    \hline
    \addlinespace[1mm]
    The e-commerce free \underline{zone} is situated in north Dubai. & district & 1 & 90\% \\
    She will have reunions in the \underline{next} few weeks. & forthcoming & 1 & 60\% \\
    It’s \underline{very} reassuring. & extraordinarily & 0 & 80\% \\
    \bottomrule
\end{tabular}
\caption{Controversial examples of contexts, target words, and substitutes (lemmas) which have high scores under either \coinco or \sword, but not the other. We consider a score to be high if it is greater than 1 for \coinco (25.3\% of substitutes) and 50\% for \sword (24.5\% of substitutes). The contexts are simplified for readability.}
\label{tab:adequacy}
\end{table*}

\section{Data analysis}\label{sec:data_analysis}
Table~\ref{tab:summary} shows overall statistics of our benchmark.
\sword comprises a total of \num{1132} context and target word pairs (418 nouns, 442 verbs, 176 adjectives, 96 adverbs) and \num{68683} total substitutes that have been labeled (including both acceptable and unacceptable substitutes).
For brevity, we defer an analysis of annotator agreement to \Cref{sec:agreement}.

\subsection{High quality}
\label{sec:quality}
With our notion of acceptability, we first observe that $75.7$\% of the substitutes from \coinco\footnote{For this analysis, we consider \coinco's substitutes that are used and labeled under \sword.} are considered unacceptable, 
and $28.1$\% of the substitutes are even inconceivable
(receiving scores less than $50$\% and $0$\% from our human annotators).
Table~\ref{tab:adequacy} shows examples of substitutes that received relatively high scores under \coinco, yet were considered unacceptable under \sword.
With the same size as \coinco (by taking the subset of our benchmark with the highest scoring substitutes per target),
the average score of the substitutes is $50.7$\% for \sword and $34.4$\% for \coinco, resulting in \quality higher quality.
Furthermore, \sword minimizes the potential noise by having fine-grained scores to account for appropriateness (Section \ref{sec:adequacy}) as well as 
explicit inconceivable substitutes, which is useful for evaluation (Section \ref{sec:evalmetrics}).

\subsection{High coverage}
\label{sec:coverage}
We show that \sword~achieves high coverage.
Among the conceivable substitutes in \sword,
13.1\% are only in \coinco (\coincoonly),
14.4\% are common to both \coinco and the thesaurus (\bothcoincoandthesaurus), 
and 72.5\% are only from thesaurus (\thesaurusonly).
Among the acceptable substitutes in \sword,
21.5\% are from \coincoonly,
36\% are from \bothcoincoandthesaurus, 
and 42.5\% are from \thesaurusonly.
This suggests that a substantial number of substitutes are not present in \coinco.
Overall, \sword contains 21.4 conceivable and 4.1 acceptable substitutes per target word on average, 
increasing those numbers by \coverage and 1.8x over \coinco, respectively.

In addition, we find that substitutes from \coincoonly are more likely to be common words whereas substitutes from \thesaurusonly are more likely to be rare words.
We compute the Zipf frequency~\citep{speer-etal-2018-wordfreq} of each substitute based on the Google $n$-gram corpus \citep{brants-franz-2006-web} and threshold conceivable substitutes into three groups: uncommon ($\leq 3.5$), neutral, common ($> 4.5$).
We observe that substitutes from \coincoonly are more likely to be common words (53.1\%) than those from \thesaurusonly (38.6\%).
On the other hand, the substitutes from \thesaurusonly tend to be more uncommon words (28.2\%) than those from \coincoonly (17.6\%).

\subsection{Reflection of appropriateness in scores}
\label{sec:adequacy}
We show that scores in \sword better reflect the appropriateness of each substitute compared to \coinco both quantitatively and qualitatively.
We consider a \coinco's score to be high if it is greater than 1 (25.3\% of substitutes) and a \sword's score to be high if it is greater than 50\% (24.5\% of substitutes).
We find that if a substitute has a high score under \coinco, it is likely to be acceptable under \sword almost half of the time (47.2\%). 
However, the converse does not hold: the acceptable substitutes under \sword have high
scores under \coinco only 29.3\% of the time. 
Intuitively, this is because \coinco's scores reflect the ease of producing the substitute
from memory, whereas \sword's scores reflect the appropriateness of the substitute. 
Table~\ref{tab:adequacy} shows examples of context, target word, and substitute triples which received high scores under either \coinco or \sword, but did not under the other benchmark.

\begin{figure}[t]
    \centering
    \vspace{-0.5cm}
    \includegraphics[width=1\linewidth]{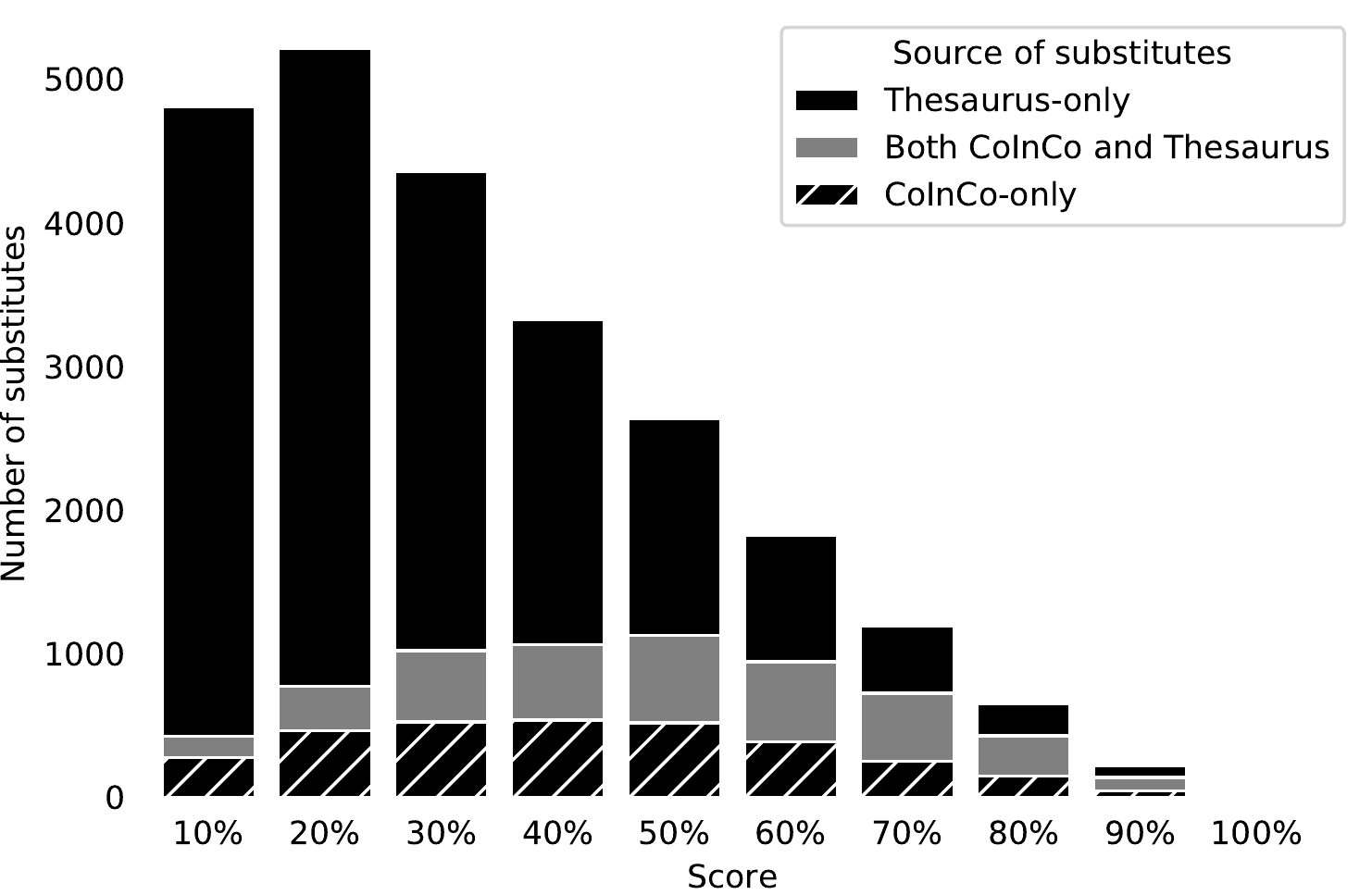}
    \caption{Score distribution of \sword's substitutes with the source of substitutes.
    We find that neither \coinco nor the thesaurus completely dominates substitutions across scores, indicating the necessity of both human-generated substitutes as well as substitutes from the thesaurus.
    Substitutes with score 0\% are not shown to make the bars visually distinguishable.\protect\footnotemark
    }
    \label{fig:source}
\end{figure}

\subsection{Validation with additional data}
\label{sec:additionaldata}
We show that the probability of an acceptable substitute falsely filtered out in Step~2 is very low.
To this end, we collected $10$ additional labels on 89 context-target word pairs randomly selected from the test set, without reducing the pool of substitutes as in Step~2.
By comparing the first three labels to the entire $10$ labels, we find that $33.7$\% of substitutes without any positive labels in Step~2 could have received one or more positive labels if they were kept in Step~3. 
However, we find that $99.2$\% of these substitutes were eventually considered unacceptable (judged by $10$ labels), indicating that the probability of an acceptable substitute incorrectly filtered out in Step~2 is very low ($0.8$\%).

\subsection{Score distribution}
\label{sec:source}
\footnotetext{1691 substitutes from \coincoonly, 917 from \bothcoincoandthesaurus, and 41817 from \thesaurusonly received a score of 0\%.}

Figure~\ref{fig:source} shows the score distribution of substitutes in \sword along with the source of substitutes: \coincoonly, \bothcoincoandthesaurus, or \thesaurusonly.
Across scores, neither \coinco nor thesaurus completely dominates substitutes, and the overlap between \coinco and thesaurus is quite small, thus indicating the necessity of both human-recalled substitutes as well as substitutes from a thesaurus.
We also find that \sword adds more substitutes for all the scores, although substitutes from the thesaurus tend to have a lower range of scores compared to those from \coinco.
Lastly, we observe that substitutes from \coinco roughly form a normal distribution, which suggests that even the substitutes provided by human annotators are controvertible, and that it is important to account for the intrinsically continuous nature of appropriateness with fine-grained scores.

\section{Model evaluation}
\label{sec:experiments}
In this section, we evaluate several methods on \sword. 
The goals of this evaluation are three-fold: 
(1) to prescribe our recommended evaluation practice for \sword,
(2) to measure performance of existing large-scale pre-trained models and state-of-the-art lexical substitution systems, and
(3) to measure human performance for the purpose of comparing current and future systems.

\subsection{Evaluation settings}
There are two primary evaluation settings in lexical substitution research: the \emph{generative} setting~\citep{mccarthy-navigli-2007-semeval} and the \emph{ranking} setting~\citep{thater-etal-2010-contextualizing}. 
In the generative setting, systems output a ranked list of substitute candidates. 
There are no restrictions on the number of candidates that a system may output. 
In the ranking setting, systems are given all substitute candidates from the benchmark (including those marked as inconceivable) and tasked with ranking them by appropriateness.
Here, we primarily focus on the generative setting and defer our experiments on the ranking setting to~\Cref{sec:app_ranking}.

\subsection{Evaluation metrics}
\label{sec:evalmetrics}
In a writing assistance context, we envision that lexical substitution systems would be used to suggest a limited number of substitutes to users (e.g. $10$ substitutes as opposed to $100$).
Hence, we consider evaluation metrics that examine the quality and coverage of the top-ranked substitutes from a system with respect to the substitutes that humans judged as \emph{acceptable} (score $> 50$\%).
Specifically, we compute precision (\pmk) and recall (\rmk) at $k$\footnote{Note that our definition of recall at $k$ is non-standard; the $\min$ compensates for the fact that there are often fewer than $k$ acceptable substitutes.}: 
\begin{align*}
P^{k} &= \frac{\text{\# acceptable substitutes in system top-}k}{\text{\# substitutes in system top-}k} \\
R^{k} &= \frac{\text{\# acceptable substitutes in system top-}k}{\min(k, \text{\# acceptable substitutes})}
\end{align*}
Because we care about both quality (precision) and coverage (recall) when comparing systems, we report \fmk{}, the harmonic mean of \pmk{} and \rmk.
Likewise, we evaluate against the list of substitutes which humans judged as \emph{conceivable} (score~$>0$\%). 
\pk{} and \rk{} constitute precision and recall of systems against this larger candidate list, and \fk{} their harmonic mean.
Motivated by past work~\citep{mccarthy-navigli-2007-semeval}, we primarily examine performance for $k=10$ and lemmatize system-generated substitutes before comparison.\footnote{Note that some methods use lemmas of target words (e.g. \bathesaurus) and others use original word forms of target words (e.g. \bagpt). We provide both forms in the benchmark.}

We note that these metrics represent a departure from standard lexical substitution methodology, established by~\citet{mccarthy-navigli-2007-semeval}. 
Like \pmk{} and \rmk{}, the previously-used \textsc{best} and \textsc{oot} metrics are also measures of precision and recall, but are not designed to take advantage of inconceivable substitutes as no such explicit negative substitutes existed in the earlier benchmarks.
Nevertheless, we report performance of all systems on these metrics in~\Cref{sec:quant_verbose} as reference.

\subsection{Baselines}
\looseness-1 We evaluate both state-of-the-art lexical substitution systems and large-scale pre-trained models as baselines on \sword. 
We reimplement the BERT-based lexical substitution system (\babbls) from~\citet{zhou-etal-2019-bert}, which achieves state-of-the-art results on past benchmarks.
As another lexical substitution system, we examine \bawordtune~\citep{ai21-2020-wordtune}, 
a commercial system which offers lexical substitution capabilities.\footnote{\bawordtune{} is not optimized for lexical substitution.}

We also examine two large-scale pre-trained models adapted to the task of lexical substitution: \babert~\citep{devlin-etal-2019-bert} and \bagpt~\citep{brown-etal-2020-gpt3}. 
To generate and rank substitute candidates with \babert, we feed in a context with a target word either masked (\babertm) or kept intact (\babertk), and output the top $50$ most likely words according to the masked language modeling head.
Because the target word is removed, \babertm{} is expected to perform poorly---its main purpose is to assess the relative importance of the presence of the target word compared to the context.
Note that both of these strategies for using \babert{} to generate candidates differ from that of \babbls, which applies dropout to the target word embedding to partially obscure it.
To generate candidates with GPT-3, 
we formulate lexical substitution as natural language generation (see \Cref{sec:app_gpt3} for details).

\begin{table}[t]
\centering
\small

\begin{tabular}{lcccc}

\toprule
& \multicolumn{2}{c}{\Gperm} & \multicolumn{2}{c}{\Gstrict} \\
\cmidrule(lr){2-3} \cmidrule(lr){4-5}
%Strategy & \ften & \fmten & \ften & \fmten \\
Model & \fmten & \ften & \fmten & \ften \\
\midrule

% https://colab.research.google.com/drive/1YL5_lnjg-PMEP8YuvlOZLMb4hw_bxkJi?usp=sharing

% https://worksheets.codalab.org/bundles/0x890c83e8b3bd40f98d3263b3bae246d9
% https://worksheets.codalab.org/bundles/0xe2a3897458f143f49261dbc196730bed
\bahumans* & $48.8$ & $77.9$ & $-$ & $-$ \\

% https://worksheets.codalab.org/bundles/0xbf4759499af340fba341bd37643f90e6
\bacoinco & $34.1$ & $63.6$ & $-$ & $-$ \\

% https://worksheets.codalab.org/bundles/0xfef43e25fa9041a6a4129f85138b2354
\bathesaurus$^\dagger$ & $25.6$ & $61.6$ & $-$ & $-$ \\

% https://worksheets.codalab.org/bundles/0x7b1dd45f890e46b7b8548e93480ef03b
\bathesaurus & $12.0$ & $44.9$ & $-$ & $-$ \\

\midrule

% https://worksheets.codalab.org/bundles/0xd1c11ab99b01414e9a73ee0d3c822bea
\bagpt & $\textbf{34.6}$ & $49.0$ & $22.7$ & $\textbf{36.3}$ \\

% https://worksheets.codalab.org/bundles/0x394156276df646bab1911b224f550605
\bawordtune$^\dagger$ & $34.6$ & $45.4$ & $\textbf{23.5}$ & $34.7$ \\

% https://worksheets.codalab.org/bundles/0x3108d95472264f8291fe35181ef55487
\bagpt$^\dagger$ & $34.4$ & $49.0$ & $22.3$ & $34.7$ \\

% https://worksheets.codalab.org/bundles/0x16cb252935ba4a0a94fc1875bdf8b7e5
\bawordtune & $34.3$ & $45.2$ & $22.8$ & $33.6$ \\

% https://worksheets.codalab.org/bundles/0x949e51794ed5460f944bc9622194ec73
\babertk$^\dagger$ & $32.4$ & $\textbf{55.4}$ & $19.2$ & $30.3$ \\

% https://worksheets.codalab.org/bundles/0x1ab3af73d9cc4ce78807c1d901206acd
\babbls & $32.1$ & $54.9$ & $17.2$ & $27.0$ \\

% https://worksheets.codalab.org/bundles/0xa2d25b46e40a48f5b0ac18737ca65e4f
\babertk & $31.7$ & $54.8$ & $15.7$ & $24.5$ \\

% https://worksheets.codalab.org/bundles/0x9b3cf951fb0849e6b1b251d6c8cb2690
\babertm & $30.9$ & $48.1$ & $10.7$ & $16.5$ \\

% https://worksheets.codalab.org/bundles/0xe827bbf485f54855b183ff543fc505e5
\babertm$^\dagger$ & $30.9$ & $48.3$ & $16.2$ & $25.4$ \\

\bottomrule

\end{tabular}
\caption{
Evaluation of systems on \sword in the generative setting.
Here, systems must both generate and rank a set of substitutes.
We observe that the performance of all baselines on all metrics falls short of human performance.
For the ``\gperm'' setting, we filter out system generated substitutes which are not in \sword.
*Computed on a subset of the test data.
$^{\dagger}$Reranked by our best ranking model (\babert).
}
\label{tab:generative}
\end{table}
\begin{table*}[t]
\centering
\small
\vspace{-0.2cm}
\begin{tabular}{p{0.2\linewidth}p{0.75\linewidth}}
    \toprule
    
    Context & The e-commerce free zone is situated in north Dubai, near the industrial free \underline{zone} in Hebel Ali \\
    
    \addlinespace[1mm]
    
    Substitutes in \sword & \textbf{sector} (90\%), \textbf{district} (90\%), \textbf{area} (90\%), \textbf{region} (70\%), \textbf{section} (70\%), \textbf{range} (60\%), \textbf{strip} (60\%), ground (50\%), segment (50\%), territory (50\%), sphere (40\%), realm (40\%), place (30\%), tract (30\%), city (30\%), belt (20\%), circuit (20\%), band (0\%) \\
    
    \midrule
    
    Reference for \fmk{} ($7$) & \textbf{sector}, \textbf{district}, \textbf{area}, \textbf{region}, \textbf{section}, \textbf{range}, \textbf{strip} \\
    
    Reference for \fk{} ($17$) & \textbf{sector}, \textbf{district}, \textbf{area}, \textbf{region}, \textbf{section}, \textbf{range}, \textbf{strip}, ground, segment, territory \\
    
    \midrule
    
    \bacoinco{} ($9$) & \textbf{area}, \textbf{region}, \textbf{district}, \textbf{section}, city, place, \textbf{range}, \textbf{strip}, territory \\
    
    \bathesaurus$^{\dagger}$ ($14$) & \textbf{district}, \textbf{area}, belt, territory, \textbf{region}, realm, \textbf{sector}, \textbf{section}, circuit, segment \\
    
    \midrule
    
    \bawordtune$^{\dagger}$ ($11$) & \textbf{district}, \textbf{area}, city, \textbf{region}, site, league, center, system, place, zona \\
    
    \bagpt$^{\dagger}$ ($13$) & \textbf{district}, \textbf{area}, territory, \textbf{region}, realm, \textbf{sector}, locality, \textbf{section}, quarter, precinct \\
    
    \babbls{} ($50$) & belt, \textbf{district}, port, \textbf{area}, zones, city, park, center, \textbf{strip}, \textbf{sector} \\
    
    \babertk$^{\dagger}$ ($50$) & zones, \textbf{district}, \textbf{area}, city, belt, \textbf{region}, park, ville, site, \textbf{sector} \\
    
    \babertm$^{\dagger}$ ($50$) & zones, \textbf{district}, \textbf{area}, city, belt, territory, \textbf{region}, haven, park, site \\
    
    \bottomrule
\end{tabular}
\caption{
Qualitative comparison of top 10 candidates generated by best systems. 
From top to bottom, table sections show
(1)~a context, \underline{target word}, substitutes, and scores from \sword (dev),
(2)~reference lists used to compute \fmk{} and \fk{} (applying thresholds of $>50\%$ and $>0\%$ to scores),
(3)~substitute candidates from oracle systems (sources of substitutes in \sword), and
(4)~substitute candidates from best systems.
For each system, we include the total number of substitute candidates produced by the system in parentheses (after removing duplicates). 
Substitutes with scores greater than 50\% from \sword{} are \textbf{bolded}.}\label{tab:generative_qualitative_succinct}
\vspace{-0.2cm}
\end{table*}

\subsection{Human and oracle systems}
\looseness-1 Here we consider human and oracle ``systems'' to help contextualize the performance of automatic lexical substitution systems evaluated on \sword.
We evaluate the performance of \bahumans{} using labels from a separate pool of annotators as described in~\Cref{sec:additionaldata}. 
Because this task is inherently subjective, this system represents the agreement of two independent sets of humans on this task, which can be viewed as a realistic upper bound for all metrics. 
We consider the substitutes that have score~$>0\%$ from the separate pool of annotators as \bahumans{}'s substitutes in the generative setting.

We also consider both of the candidate sources, \bacoinco{} and \bathesaurus{}, as oracle systems. 
Each source contains a list of substitutes for every target word,
and therefore can be viewed as a lexical substitution system and evaluated on \sword.
\bacoinco{} provides substitutes for a target word that were provided by six human annotators.
This can be thought of as a proxy for how humans perform on lexical substitution when recalling words off the top of their heads (as opposed to making binary judgements as in \bahumans{}).
\bathesaurus~provides context-free substitutes for a target word (regardless of their word senses) with the default ranking retrieved from the thesaurus.
This represents the context-insensitive ordering that a user of the same thesaurus would encounter. 

Because these oracle systems only produce candidates which are guaranteed to be in \sword, they have an inherent advantage on the evaluation metrics over other systems.
Hence, to be more equitable to other systems, we additionally compute \fmten{} and \ften{} in a ``\gperm'' fashion---filtering out model generated substitutes which are not scored under \sword (we refer to the setup without filtering as ``strict'').
It is our intention that future systems should \emph{not} use \bacoinco{} or \bathesaurus{} in any way, as they leak information about the \sword{} benchmark.

\subsection{Evaluation results}
\Cref{tab:generative} shows that the performance of all methods falls short of that of humans on all metrics. 
We interpret this as evidence that \sword{} is a challenging benchmark, since strong (albeit unsupervised) baselines like \babert{} and \bagpt{} do not reach parity with humans. 
We also observe that two models (\bawordtune{} and \bagpt) achieve higher \fmten{} than \bacoinco. 
In other words, while all models perform worse than humans who are judging the appropriateness of substitutes (\bahumans), some models appear to slightly outperform humans who are thinking of substitutes off the top of their head (\bacoinco).
This implies that some lexical substitution models may already be helpful to humans for writing assistance, 
with room for improvement. 

Overall, we find that there is no single system which emerges as the best on all metrics.
We note that, despite \babbls{} representing the state-of-the-art for past lexical substitution benchmarks, 
its performance falls short of that of commercial systems like \bagpt{} and \bawordtune{} on most criteria. 
Also, the \babert-based methods output around $5$x as many candidates as the other models on average, thus having an inherent advantage in recall with the lenient criteria (see \Cref{tab:generative_verbose} in \Cref{sec:quant_verbose}).

In \Cref{tab:generative}, we additionally report the performance of generative models by re-ranking their lists of substitutes using the best ranker from our candidate ranking evaluation, \babert{} (see \Cref{sec:app_ranking} for details).
This procedure unilaterally improves performance for all systems on all metrics except for \bagpt. % \footnote{We speculate that this is because GPT-3 produces many substitutes containing multiple word pieces, and mean-pooling several word pieces may result in lower-quality scores.}
Hence, we speculate that improved performance on the ranking setting will be mostly complementary to improved performance on the generative setting.

From a qualitative perspective, many of the systems we evaluate already produce helpful substitutes (\Cref{tab:generative_qualitative_succinct}).
In examining errors, we find that \babert-based models and \bawordtune{} tend to produce words that differ semantically from the target (e.g. ``league'' for \underline{zone}). 
Substitutes generated by \bagpt{} are often repetitive (e.g. for \underline{zone}, \bagpt{} produced $64$ substitutes, out of which only $13$ were unique)---we filter out duplicates for evaluation.
Finally, we observe that some systems produce appropriate substitutes which are not present in \sword (e.g. \bagpt{} produces ``precinct'' for \underline{zone}), 
indicating that \sword has limitations in coverage.
However, the higher coverage and quality in \sword{} compared to past benchmarks still improves the reliability of our proposed evaluation.

\section{Related work}
As we discussed previous benchmarks for lexical substitution in Section~\ref{sec:background} and relevant models in Section~\ref{sec:experiments}, we use this section to draw connections to other related literature.

\paragraph{Word sense disambiguation.}
The task of word sense disambiguation consists of selecting the intended meaning (i.e. sense) from the pre-defined set of senses for that word in a sense inventory.
The task of lexical substitution is closely related to word sense disambiguation, as many words are \emph{sense synonyms}---some of their senses are synonymous, but others are not \citep{murphy-2010-lexical}.
In fact, \citet{mccarthy-2002-lexical} proposed lexical substitution as an application-oriented word sense disambiguation task that avoids some of the drawbacks of standard word sense disambiguation, such as biases created by the choice of sense inventory \citep{kilgarriff-1997-wordsense}.

\paragraph{Near-synonym lexical choice.}
Words are often \emph{near-synonyms}---they can substitute for each other in some contexts, but not every context \citep{dimarco-etal-1993-semantic,murphy-2010-lexical}.
\sword can be viewed as a collection of human judgments on when certain near-synonyms are substitutable in a given context.
The task of near-synonym lexical choice consists of selecting the original target word from a set of candidate words given a context where the target word is masked out \citep{edmonds-hirst-2002-near}.
The candidate words are composed of the target word and its near-synonyms which are often retrieved from a lexical resource such as \citet{hayakawa-1994-ctrw}.
In this task, systems are tested whether they can reason about near-synonyms and choose the best substitute that fits in the context, without knowing any direct semantic information about the target word and without having to explicitly judge the appropriateness of other candidates.

\paragraph{Lexical and phrasal resources.}
Lexical resources such as thesauri are often used to identify possible word substitutes.
WordNet \citep{fellbaum-1998-wordnet} is a widely used lexical resource for English that includes synonymy, antonymy, hypernymy, and other relations between words.
PPDB \citep{pavlick-etal-2015-ppdb} includes both word-level and phrase-level paraphrase rules ranked by paraphrase quality.
These resources relate words and phrases in the absence of context, whereas lexical substitution requires suggesting appropriate words in context.

\paragraph{Paraphrase generation.}
Work on sentence-level paraphrase generation considers a wide range of meaning-preserving sentence transformations, including phrase-level substitutions and large syntactic changes \citep{madnani-dorr-2010-generating,wieting-gimpel-2018-paranmt,iyyer-etal-2018-adversarial,hu-etal-2019-parabank}.
Our work could be extended to phrases given appropriate methods for identifying target phrases and proposing candidate substitute phrases.
One benefit of focusing on word substitutions is that we can cover a large fraction of all appropriate substitutes,
and thus estimate recall of generative systems.
Some word-level substitutions, such as function word variation and substitutions that rely on external knowledge, are also outside the scope of our work but occur in standard paraphrase datasets \citep{bhagat-hovy-2013-squibs}.

\paragraph{Self-supervised pre-trained models.}
The task of suggesting words given surrounding context bears strong resemblance to masked language modeling,
which is commonly used for pretraining \citep{devlin-etal-2019-bert}.
However, for lexical substitution, appropriate substitutes must not only fit in context but also preserve the meaning of the target word; thus, additional work is required to make BERT perform lexical substitution \citep{zhou-etal-2019-bert,arefyev-etal-2020-comparative}.

\paragraph{Modeling human disagreement.}
In \sword, we find considerable subjectivity between annotators on the appropriateness of substitutes.
For the task of natural language inference, recent work argues that inherent disagreement between human annotators captures important uncertainty in human language processing that current NLP systems model poorly \citep{pavlick-kwiatkowski-2019-inherent,nie-etal-2020-learn}.
We hope that the fine-grained scores in \sword{} encourage the development of systems that more accurately capture the graded nature of lexical substitution.

\section*{Acknowledgments}
We sincerely thank Frieda Rong, Nelson Liu, Stephen Mussmann, Kyle Mahowald, Daniel Jiang, and all reviewers for their help and feedback throughout this project.
We also thank OpenAI and Wordtune for allowing us to evaluate their systems. 
This work was funded by DARPA CwC under ARO prime contract no. W911NF-15-1-0462.

% \todo{fix reference (inconsistent titling of proceedings (``In Proceedings of Conferencename'' vs. ``In Conferencename''), missing page numbers (e.g., line 699), missing issue numbers (e.g., line 691), and missing book/proceedings titles (e.g., line 750))}

\bibliography{anthology, custom}
\bibliographystyle{acl_natbib}

\clearpage
\appendix

\section{Data collection}

\subsection{Deduplicating context}\label{sec:dedup_context}
\sword uses the same contexts as \coinco, but with slight modifications to avoid duplication issues and incomplete contexts found in \coinco.
\coinco uses a subset of contexts from the Manually Annotated Sub-Corpus (MASC) \citep{ide-etal-2008-masc,ide-etal-2010-manually}, in which some sentences are erroneously repeated multiple times due to multiple IDs assigned to a single sentence.
Consequently, \coinco contains duplicate sentences in some contexts, as shown below:
\begin{quote}
" --was kindly received," "But an artist who would stay first among his fellows can tell when he begins to \underline{fail}." {\color{red} "But an artist who would stay first among his fellows can tell when he begins to fail."}
\end{quote}

Furthermore, we found that some parts of the document context are missing in \coinco because an ID was not assigned to the parts in MASC (e.g. ``he said.'' is missing from the above passage after the word ``received'').

To address this issue, we re-extracted full contexts from MASC.
Given a sentence containing a target word in \coinco, we located the sentence in MASC and used three non-overlapping adjacent MASC regions as our context.
As a result, our context contains additional text that was erroneously omitted in \coinco (including newlines), thereby reducing annotator confusion.
The context of the above example in our benchmark is as follows:
\begin{quote}
 " --was kindly received," {\color{blue} he said.} "But an artist who would stay first among his fellows can tell when he begins to \underline{fail}." 
{\color{blue} "Oh?"}
\end{quote}

\subsection{Retrieving substitutes from a thesaurus}\label{sec:app_thesaurus}
We use Roget's Thesaurus \citep{kipfer-2013-roget}, upon which \url{thesaurus.com} is built, as a primary source of context-free substitutes for target words in \sword.
To select substitutes for a particular target word, 
we gather substitutes from all word senses that have the same part of speech as the original target,
in order to disentangle lexical substitution from the task of word sense disambiguation as well as to include challenging distractors for evaluating models.\footnote{In \coinco, contexts, target words, and their part-of-speech tags all come from the Manually Annotated Sub-Corpus \citep{ide-etal-2008-masc,ide-etal-2010-manually}}
We use the default ranking retrieved from the thesaurus. 
When a target word has multiple word senses, 
we rank substitutes by simply concatenating the lists of substitutes for each sense in the order that the senses appear in the thesaurus.

\section{Crowdsourcing}
\label{app:crowdsourcing}

\subsection{Instructions and interface}
\begin{figure*}[t]
    \centering
    \includegraphics[width=1\linewidth]{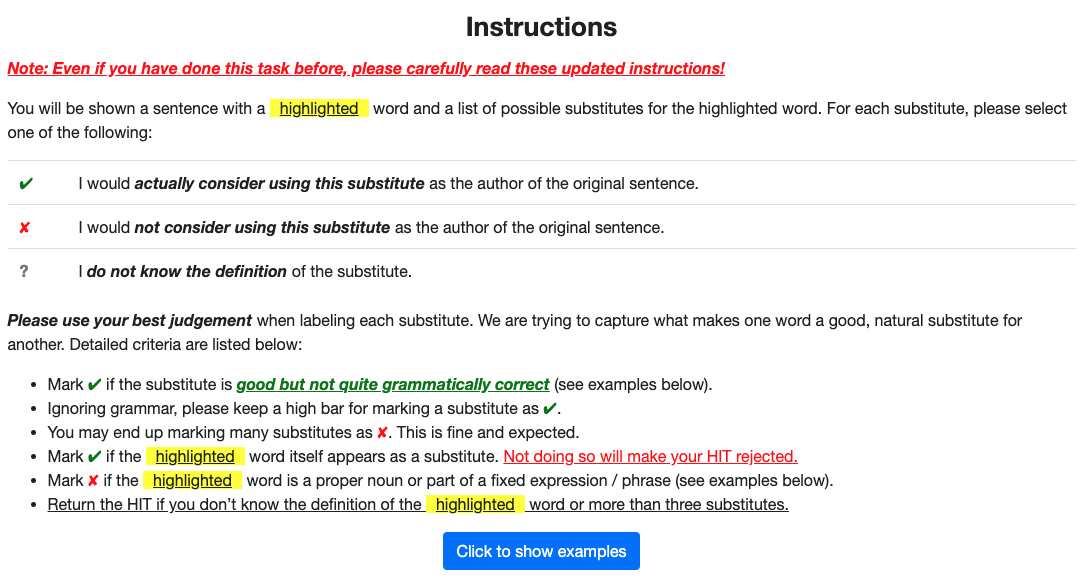}
    \includegraphics[width=1\linewidth]{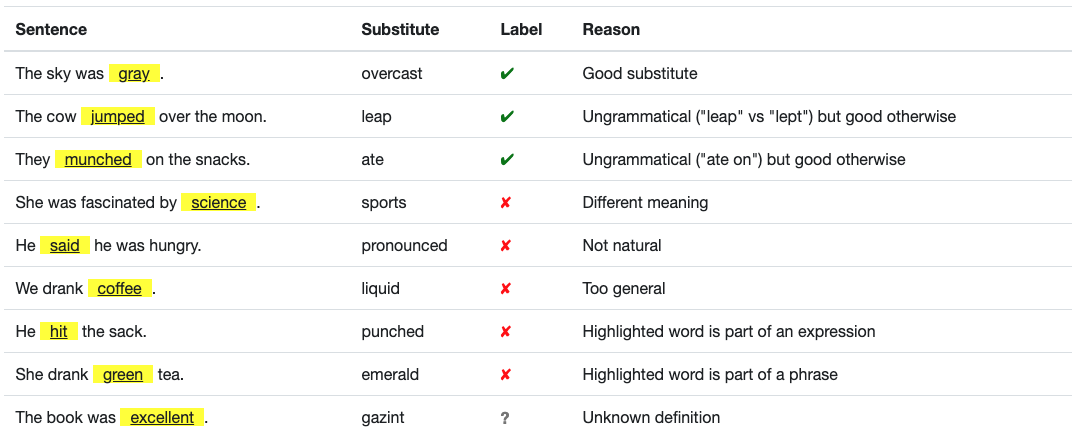}
    \caption{Our instructions ask annotators to accept a substitute if they would actually consider using the substitute as the author of the context, reject if not, or abstain if they do not know the definition of the substitutes.
    Examples are provided for reference, when annotators click the ``click to show examples'' buttons.}
    \label{fig:instructions}
\end{figure*}

\begin{figure*}[t]
    \centering
    \includegraphics[width=1\linewidth]{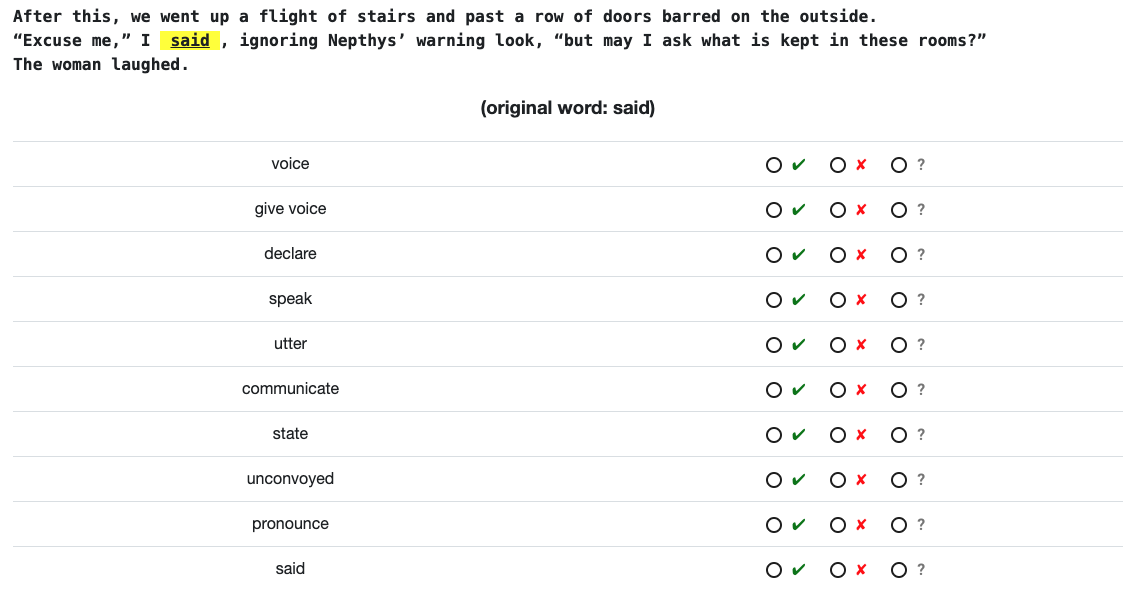}
    \caption{Our Human Intelligence Task (HIT) contains a highlighted target word in the context, which consisted of three sentences to provide sufficient context. 
    For each HIT, we show at most 10 candidate substitutes for the target word and two control substitutes for filtering.}
    \label{fig:interface}
\end{figure*}

Figures~\ref{fig:instructions} and \ref{fig:interface} show the instructions and interface we used for Amazon Mechanical Turk (AMT) to crowdsource labels on substitutes.
Following the practice of \coinco, we showed a highlighted target word in the context, which consisted of three sentences to provide sufficient context.
We instructed annotators to provide a negative label if the target word is a proper noun or part of a fixed expression or phrase.

Since our Human Intelligence Task (HIT) concerns acceptability judgement as opposed to substitute generation, we made the following modifications to the \coinco's setup.
First, we asked annotators whether they ``would actually consider using this substitute'' rather than whether the substitute ``would not change the meaning'' of the target word (Section \ref{sec:addressing}).
Second, we allowed annotators to abstain if they do not know the definition of the substitute, while asking them to return the HIT if they do not know the definition of the target word or more than three substitutes.
Third, we asked annotators to accept a substitute which is ``good but not quite grammatically correct.''
Lastly, we asked annotators to accept the substitute identical to the target word, in attempt to filtering out spammed HITs (Section~\ref{sec:filtering}).

\subsection{Setting on Amazon Mechanical Turk}
Each HIT contained at most 10 candidate substitutes for a context-target word pair.
When there were more than 10 candidate substitutes, we generated multiple HITs by partitioning the candidate substitutes into multiple subsets with potentially different length, using $\mathtt{numpy.array\_split}$.
We randomized the ordering of substitutes so that each HIT is likely to contain substitutes from both \coinco and the thesaurus.
The following qualification conditions were used to allow experienced annotators to participate in our task:
\begin{itemize}
    \item HIT Approval Rate (\%) for all Requesters’ HITs is greater than 98.
    \vspace{-0.2cm}\item Location is the United States.
    \vspace{-0.2cm}\item Number of HITs Approved is greater than 10,000.
\end{itemize}

Our target hourly wage for annotators was \$15. 
Based on our in-person pilot study with five native English speakers, we approximated the time per assignment (labeling at most twelve substitutes) to be 25 seconds. 
Then, we assumed that it may take 1-2x longer for crowd workers to complete the assignments and decided on the compensation of \$0.10 to fall into the range of \$7.25 (US federal minimum wage) and \$15 per hour, which corresponds to 50 seconds and 24 seconds per assignment, respectively.

It may be surprising that our assignments only take~25 seconds on average, though there are several reasons why this is the case: (1) In general, making binary judgements about substitute words takes very little time for native speakers. (2) Annotators only have to read the target sentence once to provide judgements for all substitutes in an assignment. (3) Annotators usually do not need to read the two additional context sentences to make judgements. (4) Annotators can almost instantly judge two control substitutes (Section \ref{sec:filtering}), and are therefore only realistically evaluating at most ten candidates per assignment.

\subsection{Filtering spam}
\label{sec:filtering}

In order to filter out work done by spammers, we included two additional control candidate substitutes in every HIT: 
the original target word and a randomly chosen dictionary word.
Annotators were instructed to accept the substitute identical to the target word and were expected to either reject or abstain on the random word.
We used these control substitutes to filter out spammed HITs.
Concretely, we filtered out all the HITs with any wrong label assigned to the control substitutes as well as HITs completed by annotators whose overall accuracy on control substitutes across HITs was less than $90$\%.
Then, we re-collected labels on these filtered HITs for Step~2 and Step~3.

\section{Data analysis}

\subsection{Annotator agreement}
\label{sec:agreement}
\begin{figure}[t]
    \centering
    \includegraphics[width=1\linewidth]{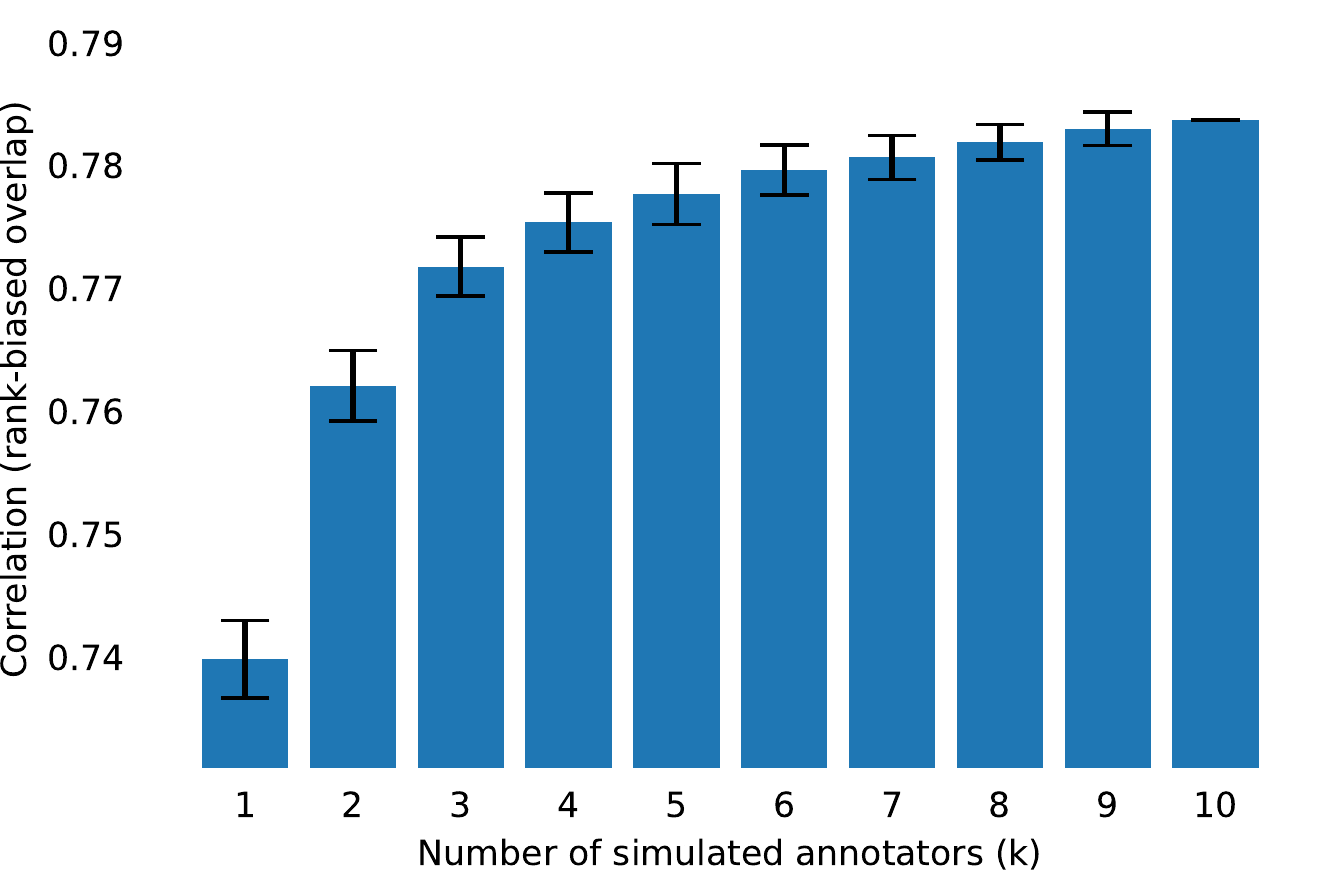}
    \caption{Annotator agreement between \sword and $k$ additional annotators measured by rank-biased overlap (RBO) \citep{webber-etal-2010-similarity}.
    The standard deviations from $100$ simulations are shown as error bars.
    We observe quite low RBO for $k < 3$ and diminishing returns as $k$ grows.
    This indicates that there is wide variation in opinions, and it is necessary to use sufficiently large $k$ to capture the distribution.
    }
    \label{fig:agreement}
\end{figure}

\citet{mccarthy-navigli-2007-semeval} introduced two inter-annotator agreement measures, which assumes that a fixed number of annotators generate a set of substitutes for every context-target word pair.\footnote{\citet{mccarthy-navigli-2007-semeval} had five annotators from the UK, where each of them annotated the entire dataset.}
However, these measures are not designed for the case when there is only one collective set of substitutes for each context-target word pair, and every context-target word pair is labeled by various combinations of annotators.

Instead, we compute correlation between the two ranked lists using rank-biased overlap (RBO) \citep{webber-etal-2010-similarity}, which handles non-conjoint lists and weights high ranks more heavily than low unlike other common rank similarity measures such as Kendall’s $\tau$ and Spearman’s $\rho$.
With additionally collected 10 labels (Section~\ref{sec:additionaldata}), we computed RBO by comparing the ranked list of substitutes derived from the data to that of \sword~and simulate the effect of having $k$ annotators by sampling $k$ labels per substitute without replacement a total of $100$ times.

Figure~\ref{fig:agreement} shows the correlation between \sword and $k$ additional human annotators.
We observe quite low RBO for $k < 3$ and diminishing returns as $k$ grows.
Based on this observation, we argue that there is wide variation in opinions and it is necessary to use sufficiently large $k$ to capture the distribution.

\section{Model evaluation}
\label{sec:app_ranking}

\subsection{Ranking setting}
\begin{table}[t]
\centering
%\small

\begin{tabular}{lcc}

% >>>>>>>>>>>>>>>> BEGIN PASTE

\toprule
Model & GAP \\
\midrule

% https://colab.research.google.com/drive/1YL5_lnjg-PMEP8YuvlOZLMb4hw_bxkJi?usp=sharing

% https://worksheets.codalab.org/bundles/0x98688dcfdb384e9891045db93e4572f5
\bahumans* & $66.2$ \\

\midrule

% https://worksheets.codalab.org/bundles/0x4560de65200846ca81af34e839febfc6
\babert & $\textbf{56.9}$ \\

% https://worksheets.codalab.org/bundles/0x8f79ecda177c44c29d4991b6b87cadca
\babbls & $53.4$ \\

% https://worksheets.codalab.org/bundles/0xc09a2aa624264b6c8eb7114d96210173
\babbls{} w/o $s_p$ & $52.9$ \\

% https://worksheets.codalab.org/bundles/0x1e1a6af9025c4880b643a593a4ac383e
\babbls{} w/o $s_v$ & $51.7$ \\

% https://worksheets.codalab.org/bundles/0x1cede47df3114985b309b977e16bdda6
\baglove & $49.7$ \\

% https://worksheets.codalab.org/bundles/0x2a8e097c10e84a47823073eb60e3b73a
\barandom & $32.7$ \\

\bottomrule

% <<<<<<<<<<<<<<<< END PASTE

\end{tabular}
\caption{Evaluation of models on \sword in the ranking setting. 
Here, systems provide a score for every candidate in the benchmark. *Computed on a subset of the test data.}
\label{tab:ranking}
\end{table}
As opposed to the generative setting where systems must generate and rank substitute candidates, 
in the (easier) ranking setting, 
systems are given all substitute candidates from the benchmark (including those marked as inconceivable) and tasked with ranking them by their appropriateness.

\subsection{Evaluation metrics}
To evaluate ranking models, we adopt standard practice and report generalized average precision (GAP)~\citep{kishida-2005-property}. 
GAP is similar to mean average precision, but assigns more credit for systems which produce substitutes that have higher scores in the reference list. 
Considering that our data collection procedure results in 
reference scores which correspond more to substitute appropriateness than ease-of-recollection, 
GAP is aligned with our high-level goals.

\subsection{Baselines}
We evaluate contextual embeddings from \babert{} and word embeddings from \baglove~\citep{pennington-etal-2014-glove}, using cosine similarity of the target and substitute embeddings as the score. 
To compute the contextual embedding of a target or substitute with \babert{}, we mean pool contextual embeddings of its constituent word pieces. 
Because \baglove{} discards contextual information, we expect it to perform worse than \babert{}, and is mainly used to assist interpretation of GAP scores. 
In the ranking setting, we are unable to evaluate \bagpt{} and \bawordtune{}, as we interface with these systems via an API which provides limited access to the underlying models.
We report GAP scores in~\Cref{tab:ranking}.

\subsection{Results}
We posit that contextual word embedding models should be invariant to contextual synonymy---they should embed acceptable substitutes nearby to one another. 
Hence, the \sword{} ranking setting may offer a useful perspective for evaluating this aspect of such models. 
In the ranking setting, our best contextual embedding model (\babert) achieves a GAP score of $56.9$. 
While \babert{} outperforms a simple context-free baseline (\baglove), it falls short of the $66.2$ GAP score achieved by \bahumans. 
We interpret this as evidence that contextual embedding models have room to improve before attaining the aforementioned invariance.

\subsection{Lexical substitution as natural language generation}
\label{sec:app_gpt3}
GPT-3 is a language model which generates text in a left-to-right order, and is not designed specifically for the task of lexical substitution. 
To use GPT-3 to perform lexical substitution, we formulate the task in terms of natural language generation, and use in-context learning as described in~\citep{brown-etal-2020-gpt3}. 
Specifically, we draw examples at random from the \sword development set to construct triplets of text consisting of (context with target word indicated using asterisks, natural language query, comma-separated list of all substitutes with score $> 0$\% in descending score order) as follows:

\begin{quote}
Phone calls were monitored. An undercover force of Manhattan Project security agents **infiltrated** the base and bars in the little town of Wendover (population 103) to spy on airmen. Karnes knew the 509th was preparing for a special bombing mission, but he had no idea what kind of bombs were involved.

Q: What are appropriate substitutes for **infiltrated** in the above text?

A: penetrate, swarm, break into, infest, overtake, encompass, raid, breach
\end{quote}

We construct as many of these priming triplets as as can fit in GPT-3's $2048$-token context (roughly $12$ examples on average), leaving enough room for a test example formatted the same way except without the list of answers. 
Then, we query the 175B-parameter $\mathtt{davinci}$ configuration of GPT-3 to generate a text continuation with up to $128$ tokens. 
Finally, we parse the generated text from GPT-3, using its natural language ordering as the ordering for evaluation.

In an initial pilot study on a random split of our development set, we selected the sampling hyperparameters for GPT-3 as $\mathtt{temperature}$ $0$, $\mathtt{presence\_penalty}$ $0.5$, 
and $\mathtt{frequency\_penalty}$ $0$, 
among possible candidates of $\{0, 1\}$ and $\{0, 0.5, 1.0\}$, and $\{0, 0.5, 1.0\}$, respectively. 
We used a grid search ($18$ runs) to select values based on highest \ften.

\section{Additional evaluation results}\label{sec:quant_verbose}

We include additional results from our evaluation. 
In~\Cref{tab:generative_verbose}, we break down \fmten{} from~\Cref{tab:generative} into \pmten{} and \rmten. 
In~\Cref{tab:generative_traditional}, we report performance of all generative baselines on traditional metrics for lexical substitution. 

\begin{table*}[t]
\centering

\begin{tabular}{lcccccccccccc}

% Source: https://docs.google.com/spreadsheets/d/1ZC6nctw0_YIuLbw3KNHzSwaMJv6JCZVz6jTNbPBey3E/edit?usp=sharing
% Render to LaTeX: https://colab.research.google.com/drive/1TXo0g0t5mX2Fvu-FL_xpmKGm4gbQFN0e#scrollTo=EnoYDaboEv7_
% NOTE: Codalab URLs reflect source of *results file* but not necessarily source of *metrics*
% TODO: Once metrics are finalized, run eval on codalab
% TODO: Fix column ordering!!!

% >>>>>>>>>>>>>>>> BEGIN PASTE

\toprule
& \multicolumn{6}{c}{\Gperm} & \multicolumn{6}{c}{\Gstrict} \\
\cmidrule(lr){2-7} \cmidrule(lr){8-13}
Model & \fmten & \pmten & \rmten & \ften & \pten & \rten & \fmten & \pmten & \rmten & \ften & \pten & \rten \\
\midrule

% https://worksheets.codalab.org/bundles/0x890c83e8b3bd40f98d3263b3bae246d9
% https://worksheets.codalab.org/bundles/0xe2a3897458f143f49261dbc196730bed
\bahumans* & $48.8$ & $43.9$ & $54.8$ & $77.9$ & $76.7$ & $79.1$ & $-$ & $-$ & $-$ & $-$ & $-$ & $-$ \\

% https://worksheets.codalab.org/bundles/0xbf4759499af340fba341bd37643f90e6
\bacoinco & $34.1$ & $24.3$ & $57.0$ & $63.6$ & $71.5$ & $57.3$ & $-$ & $-$ & $-$ & $-$ & $-$ & $-$ \\

% https://worksheets.codalab.org/bundles/0xfef43e25fa9041a6a4129f85138b2354
\bathesaurus$^\dagger$ & $25.6$ & $17.0$ & $51.8$ & $61.6$ & $60.4$ & $62.8$ & $-$ & $-$ & $-$ & $-$ & $-$ & $-$ \\

% https://worksheets.codalab.org/bundles/0x7b1dd45f890e46b7b8548e93480ef03b
\bathesaurus & $12.0$ & $8.0$ & $24.2$ & $44.9$ & $44.1$ & $45.8$ & $-$ & $-$ & $-$ & $-$ & $-$ & $-$ \\

\midrule

% https://worksheets.codalab.org/bundles/0xd1c11ab99b01414e9a73ee0d3c822bea
\bagpt & $\textbf{34.6}$ & $29.8$ & $41.4$ & $49.0$ & $76.0$ & $36.1$ & $22.7$ & $15.9$ & $\textbf{40.0}$ & $\textbf{36.3}$ & $39.2$ & $\textbf{33.8}$ \\

% https://worksheets.codalab.org/bundles/0x394156276df646bab1911b224f550605
\bawordtune$^\dagger$ & $34.6$ & $\textbf{31.3}$ & $38.7$ & $45.4$ & $\textbf{76.5}$ & $32.3$ & $\textbf{23.5}$ & $\textbf{17.2}$ & $37.0$ & $34.7$ & $\textbf{41.0}$ & $30.1$ \\

% https://worksheets.codalab.org/bundles/0x3108d95472264f8291fe35181ef55487
\bagpt$^\dagger$ & $34.4$ & $29.6$ & $41.2$ & $49.0$ & $76.1$ & $36.2$ & $22.3$ & $15.6$ & $39.2$ & $34.7$ & $37.4$ & $32.3$ \\

% https://worksheets.codalab.org/bundles/0x16cb252935ba4a0a94fc1875bdf8b7e5
\bawordtune & $34.3$ & $31.0$ & $38.4$ & $45.2$ & $76.2$ & $32.2$ & $22.8$ & $16.7$ & $35.9$ & $33.6$ & $39.7$ & $29.1$ \\

% https://worksheets.codalab.org/bundles/0x949e51794ed5460f944bc9622194ec73
\babertk$^\dagger$ & $32.4$ & $24.4$ & $\textbf{48.2}$ & $\textbf{55.4}$ & $68.7$ & $\textbf{46.4}$ & $19.2$ & $12.7$ & $38.9$ & $30.3$ & $29.6$ & $30.9$ \\

% https://worksheets.codalab.org/bundles/0x1ab3af73d9cc4ce78807c1d901206acd
\babbls & $32.1$ & $24.2$ & $47.7$ & $54.9$ & $68.2$ & $45.9$ & $17.2$ & $11.4$ & $34.9$ & $27.0$ & $26.4$ & $27.6$ \\

% https://worksheets.codalab.org/bundles/0xa2d25b46e40a48f5b0ac18737ca65e4f
\babertk & $31.7$ & $23.8$ & $47.2$ & $54.8$ & $67.9$ & $45.9$ & $15.7$ & $10.4$ & $31.8$ & $24.5$ & $24.0$ & $25.0$ \\

% https://worksheets.codalab.org/bundles/0x9b3cf951fb0849e6b1b251d6c8cb2690
\babertm & $30.9$ & $25.6$ & $39.0$ & $48.1$ & $70.3$ & $36.6$ & $10.7$ & $7.1$ & $21.8$ & $16.5$ & $16.2$ & $16.9$ \\

% https://worksheets.codalab.org/bundles/0xe827bbf485f54855b183ff543fc505e5
\babertm$^\dagger$ & $30.9$ & $25.6$ & $39.0$ & $48.3$ & $70.6$ & $36.7$ & $16.2$ & $10.7$ & $32.8$ & $25.4$ & $24.8$ & $25.9$ \\

\bottomrule

% <<<<<<<<<<<<<<<< END PASTE

\end{tabular}
\caption{
Expansion on the results from \Cref{tab:generative}, breaking down F-measures by precision and recall.
*Computed on a subset of the test data.
$^{\dagger}$Reranked by our best ranking model (\babert).
}
\label{tab:generative_verbose}
\end{table*}
\begin{table*}[t]
\centering

\begin{tabular}{lccccc}

% Source: https://docs.google.com/spreadsheets/d/1ZC6nctw0_YIuLbw3KNHzSwaMJv6JCZVz6jTNbPBey3E/edit?usp=sharing
% Render to LaTeX: https://colab.research.google.com/drive/1TXo0g0t5mX2Fvu-FL_xpmKGm4gbQFN0e#scrollTo=EnoYDaboEv7_
% NOTE: Codalab URLs reflect source of *results file* but not necessarily source of *metrics*
% TODO: Once metrics are finalized, run eval on codalab

% >>>>>>>>>>>>>>>> BEGIN PASTE

\toprule
Model & \textsc{Best} & \textsc{Best-M} & \textsc{Oot} & \textsc{Oot-M} & $P^1$ \\
\midrule

% https://worksheets.codalab.org/bundles/0xdcf28e5e769a46598c25f4a1efe4d919
\baoracle & $6.7$ & $96.9$ & $71.5$ & $99.4$ & $100.0$ \\

% https://worksheets.codalab.org/bundles/0x890c83e8b3bd40f98d3263b3bae246d9
%\bahumans* & $8.1$ & $29.6$ & $25.9$ & $56.9$ & $91.0$ \\

% https://worksheets.codalab.org/bundles/0xe2a3897458f143f49261dbc196730bed
\bahumans* & $3.4$ & $29.4$ & $54.0$ & $92.2$ & $87.6$ \\

% https://worksheets.codalab.org/bundles/0xbf4759499af340fba341bd37643f90e6
\bacoinco & $5.5$ & $26.3$ & $39.2$ & $67.9$ & $88.2$ \\

% https://worksheets.codalab.org/bundles/0xfef43e25fa9041a6a4129f85138b2354
\bathesaurus$^\dagger$ & $2.1$ & $18.8$ & $37.7$ & $52.8$ & $80.4$ \\

% https://worksheets.codalab.org/bundles/0x7b1dd45f890e46b7b8548e93480ef03b
\bathesaurus & $2.1$ & $7.9$ & $24.1$ & $24.6$ & $63.0$ \\

\midrule

% https://worksheets.codalab.org/bundles/0xd1c11ab99b01414e9a73ee0d3c822bea
\bagpt & $2.5$ & $\textbf{22.4}$ & $\textbf{22.8}$ & $45.5$ & $70.9$ \\

% https://worksheets.codalab.org/bundles/0x394156276df646bab1911b224f550605
\bawordtune$^\dagger$ & $\textbf{2.7}$ & $20.8$ & $20.2$ & $41.4$ & $\textbf{73.5}$ \\

% https://worksheets.codalab.org/bundles/0x3108d95472264f8291fe35181ef55487
\bagpt$^\dagger$ & $2.5$ & $18.7$ & $21.9$ & $44.3$ & $61.4$ \\

% https://worksheets.codalab.org/bundles/0x16cb252935ba4a0a94fc1875bdf8b7e5
\bawordtune & $2.7$ & $15.7$ & $19.6$ & $40.6$ & $61.4$ \\

% https://worksheets.codalab.org/bundles/0x949e51794ed5460f944bc9622194ec73
\babertk$^\dagger$ & $0.7$ & $16.8$ & $22.4$ & $\textbf{46.4}$ & $60.0$ \\

% https://worksheets.codalab.org/bundles/0x1ab3af73d9cc4ce78807c1d901206acd
\babbls & $0.7$ & $17.8$ & $20.0$ & $42.2$ & $54.3$ \\

% https://worksheets.codalab.org/bundles/0xa2d25b46e40a48f5b0ac18737ca65e4f
\babertk & $0.7$ & $15.7$ & $18.3$ & $40.0$ & $48.3$ \\

% https://worksheets.codalab.org/bundles/0x9b3cf951fb0849e6b1b251d6c8cb2690
\babertm & $0.5$ & $7.9$ & $12.7$ & $25.9$ & $29.3$ \\

% https://worksheets.codalab.org/bundles/0xe827bbf485f54855b183ff543fc505e5
\babertm$^\dagger$ & $0.5$ & $18.0$ & $18.6$ & $37.7$ & $58.0$ \\

\bottomrule

% <<<<<<<<<<<<<<<< END PASTE

\end{tabular}
\caption{Evaluation of models on SWORD in the generative setting using traditional evaluation metrics.
We also include numbers for an \baoracle, as (unlike for \fmten{} and GAP), the oracle does not achieve a score of $100$.
*Computed on a subset of the test data. 
$^{\dagger}$Reranked by our best ranking model (\babert).
%\todo{Chris: also would be nice to have F1 score here for convenience}
% CHRIS: It would be a bit awkward to add this here because we would have to break F-measures down by lenient and strict. I think it's fine as is
}
\label{tab:generative_traditional}
\end{table*}

\end{document}